\documentclass[sigconf]{acmart}

\renewcommand\footnotetextcopyrightpermission[1]{} 
\usepackage{amsmath,amsfonts,bm}

\def\eqref#1{equation~\ref{#1}}

\def\1{\bm{1}}

\DeclareMathAlphabet{\mathsfit}{\encodingdefault}{\sfdefault}{m}{sl}
\SetMathAlphabet{\mathsfit}{bold}{\encodingdefault}{\sfdefault}{bx}{n}

\newcommand{\ourGraph}{\textsc{SNOMED-X}}

\usepackage{hyperref}
\usepackage{url}
\usepackage{graphicx}
\usepackage{tabu}
\usepackage{multirow}
\usepackage{enumitem}
\usepackage[T1]{fontenc}
\usepackage[utf8]{inputenc}
\newcommand{\bb}{\mathbf{b}}

\newcommand{\hb}{\mathbf{h}}

\newcommand{\vbb}{\mathbf{v}}

\newcommand{\xb}{\mathbf{x}}
\newcommand{\yb}{\mathbf{y}}

\acmConference[2019 KDD Workshop on Applied Data Science for Healthcare]{DSHealth '19}{Aug. 5, 2019}{Anchorage, AK}

\begin{document}

\title{Snomed2Vec: Random Walk and Poincar\'e Embeddings of a  Clinical Knowledge Base for Healthcare Analytics}


\author{Khushbu Agarwal}
\affiliation{%
  \institution{Pacific Northwest National Laboratory}
}
\email{khushbu.agarwal@pnnl.gov}

\author{Tome Eftimov}
\affiliation{%
  \institution{Stanford University}
}
\email{teftimov@stanford.edu}

\author{Raghavendra Addanki}
\affiliation{%
  \institution{University of Massachusetts Amherst}
}
\email{raddanki@cs.umass.edu}

\author{Sutanay Choudhury}
\affiliation{%
  \institution{Pacific Northwest National Laboratory}
}
\email{sutanay.choudhury@pnnl.gov}

\author{Suzanne Tamang}
\affiliation{%
  \institution{Stanford University}
}
\email{stamang@stanford.edu}

\author{Robert Rallo}
\affiliation{%
  \institution{Pacific Northwest National Laboratory}
}
\email{robert.rallo@pnnl.gov}

\begin{abstract}
Representation learning methods that transform encoded data (e.g., diagnosis and drug codes) into continuous vector spaces (i.e., vector embeddings) are critical for the application of deep learning in healthcare. Initial work in this area explored the use of variants of the word2vec algorithm to learn embeddings for medical concepts from electronic health records or medical claims datasets. We propose learning embeddings for medical concepts by using graph-based representation learning methods on SNOMED-CT, a widely popular knowledge graph in the healthcare domain with numerous operational and research applications.  Current work presents an empirical analysis of various embedding methods, including the evaluation of their performance on multiple tasks of biomedical relevance (node classification, link prediction, and patient state prediction). Our results show that concept embeddings derived from the SNOMED-CT knowledge graph significantly outperform state-of-the-art embeddings, showing 5-6x improvement in ``concept similarity" and 6-20\% improvement in patient diagnosis. 
\end{abstract}

\maketitle

\section{Introduction}

Nowadays, health informatics applications in information retrieval, question answering, diagnostic support, and predictive analytics aim at leveraging deep learning methods. Efficient approaches to learn continuous vector representations from discrete inputs (e.g., diagnostic or drug codes) are required to implement most deep learning workflows.  

\subsection{Related work}

Early work on representation learning of medical concepts focused on extending widely popular skip-gram based models to medical text corpora \cite{minarro2015} and clinical text \cite{devine2014med}.


More recently, Choi et al \cite{choi2016multi} proposed learning concept representations from datasets of longitudinal electronic health records (EHR) using temporal and co-occurrence information from patients' doctor visits. The resulting embeddings, referred to as \textsl{Med2Vec}, are available for nearly 27k ICD-9 codes.  The CUI2Vec algorithm \cite{beam2018clinical} learns embeddings from a combination of medical text, EHR datasets, and clinical notes. Concepts in each of these data sources are mapped to a single thesaurus from the Unified Medical Language System (UMLS), and a concept co-occurrence matrix is constructed from multiple data sources and used to learn the embeddings. Despite recent advances, several factors limit the use and adoption of these embeddings. For instance, most EHR or claims datasets are of limited distribution and accessibility. Without open access to the original data, it is difficult to reproduce and understand how derived embeddings were shaped by the coverage of medical concepts in the datasets as well as by algorithmic decisions. The main contributions of this work are:
\begin{enumerate}[leftmargin=*]
    \item We learn vector representations for medical concepts by using graph-embedding learning methods such as node2vec (exploiting random walk-based connectivity), metapath2vec (emphasizing multi-relational properties), and Poincar\'e (learning taxonomical hierarchies in hyperbolic space) on SNOMED-CT \cite{donnelly2006snomed}. SNOMED-CT/UMLS is a publicly available open resource that provides coverage for over 300,000 clinical concepts and facilitates interoperability and reproducibility. 
    \item Our experiments suggest that SNOMED-derived embeddings outperform Med2vec and Cui2vec on multiple machine-learning tasks that are critical for healthcare applications. 
    \item Our code and pre-trained embeddings for over 300,000 medical concepts are available as an open-source resource for downstream applications and for reproducibility of current results.\footnote{Available at \url{https://gitlab.com/agarwal.khushbu/Snomed2Vec}.}
\end{enumerate}

\begin{figure*}[htbp]
\begin{centering}
\includegraphics[width=1.2\columnwidth]{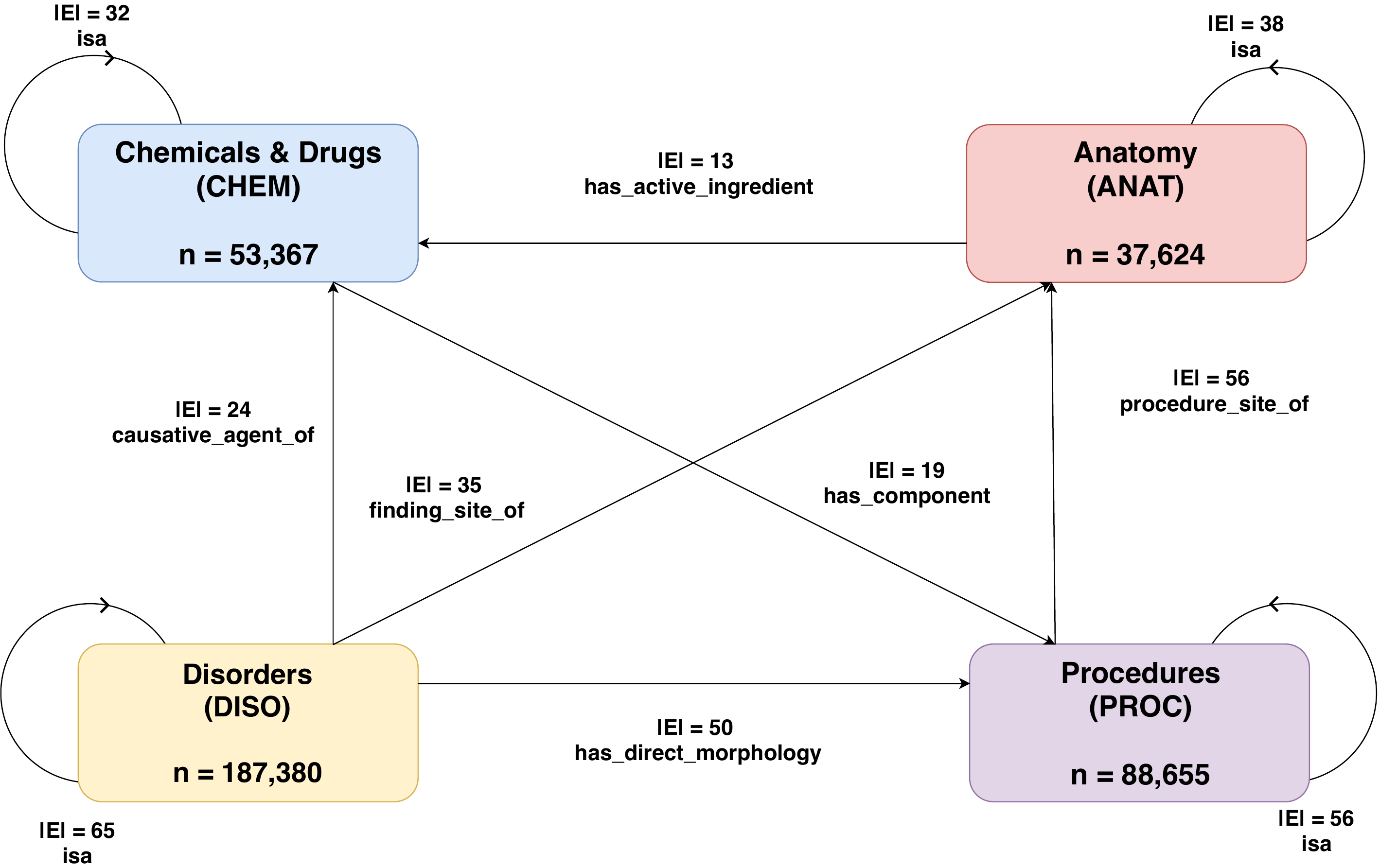}
\caption{Schema of the subset of SNOMED-CT knowledge graph extracted from UMLS. `n' denotes the number of concepts of each type and |E| indicates unique number of relationships types between the concept pairs.}
\label{fig:snomed}
\end{centering}
\vspace{-4mm}
\end{figure*}

The paper is organized as follows.  Section \ref{sec:Background} provides a quick overview of SNOMED.  Section \ref{sec:Representation learning} describes the graph-based embedding learning approaches and healthcare analytics tasks used in our empirical studies.  Sections 4 and 5 describe the experimental setup and provide conclusions from the empirical evaluation of various embedding learning methodologies.

\section{Background}
\label{sec:Background}
\subsection{SNOMED: overview and applications}
\begin{figure}[htbp]
\begin{centering}
\includegraphics[width=1.05\columnwidth]{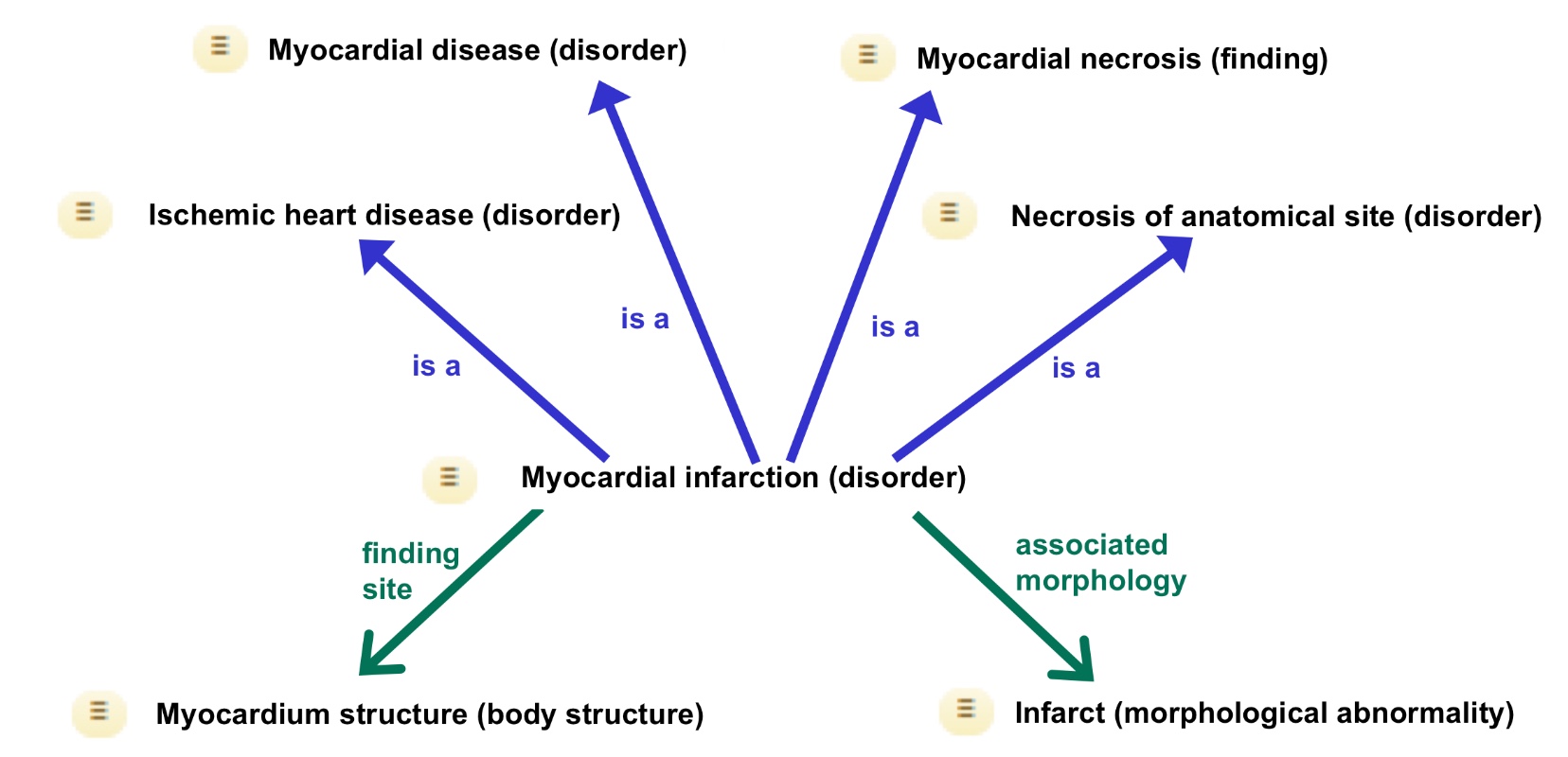}
\caption{Illustration of semantic types and relationships in SNOMED knowledge graph (source: https://confluence.ihtsdotools.org.  }
\label{fig:snomed_example}
\end{centering}
\vspace{-4mm}
\end{figure}

SNOMED Clinical Terms (CT) is the most widely used clinical terminology by healthcare researchers and practitioners in the world, providing codes, terms, synonyms and definitions used for documentation and reporting within health systems \cite{donnelly2006snomed}. It is based on UMLS, which maps terms across sixty controlled vocabularies, and hence is a critical resource for data scientists seeking to fuse heterogeneous types of observational data from the same source (e.g., drugs and disorders) or to combine disparate data sources of the same type (e.g., patient data from two discrete EHR installations) into a common ontological framework. 
Figure \ref{fig:snomed} shows a simplified view of SNOMED-CT concept model, demonstrating clinical concepts of interest and the accompanying relationships. Figure \ref{fig:snomed_example} shows the illustration of node and some of it's associated relations in the SNOMED knowledge graph.
\section{Representation learning methods and evaluation tasks}
\label{sec:Representation learning}
\subsection{Representation learning}
Given a graph $G = (V, E)$, representation learning algorithms map each node $u \in V(G)$ to a real-valued vector in a low-dimensional space $d$, such that $\Gamma(u) \in \mathbb{R}^d$, where $d << |V(G)|$.

\textbf{Embeddings in Euclidean space.} Representation learning algorithms for networks can be broadly separated into two groups based on their reliance on matrix factorization versus random walks.  Random walk-based methods, such as DeepWalk \cite{perozzi2014deepwalk} and Node2vec \cite{grover2016node2vec}, try to learn representations that roughly minimize a cross-entropy loss function of the form $\sum_{v_i, v_j \in V(G)} -log(p_L(v_j | v_i))$, where $p_L(v_j | v_i)$ is the probability of visiting a node $v_j$ on a random walk of length $L$ starting from node $v_i$.  Algorithms similar to Node2vec are further extended by Metapath2vec \cite{dong2017metapath2vec} to incorporate multi-relational properties by constraining random walks.

\textbf{Embeddings in Hyperbolic space} Many real-world datasets exhibit hierarchical structure.  Recently, hyperbolic spaces have been advocated as alternatives to the standard Euclidean spaces in order to better represent the hierarchical structure \cite{nickel2017poincare, dhingra2018embedding}.  The Poincar\'e ball model is a popular way to model a hyperbolic space within more familiar Euclidean space.  The distance between two points $u$ and $v$ within the Poincar\'e ball is given as,
\begin{equation}
    d(u, v) = cosh^{-1}(1+2\frac{||u-v||^2}{(1-||u||^2)(1-||v||^2)})
\end{equation}
Given $N(u)$ as the set of negative samples for an entity $u$, the loss function maximizing the distance between unrelated samples is,
\begin{equation}
    \zeta = \sum_{(u, v)} \log{\frac{e^{-d(u, v)}}{\sum_{v_1 \in N(u)} e^{-d(u, v_1)}}}
\end{equation}

\begin{figure*}[htbp]
\begin{centering}
\includegraphics[width=2.0\columnwidth]{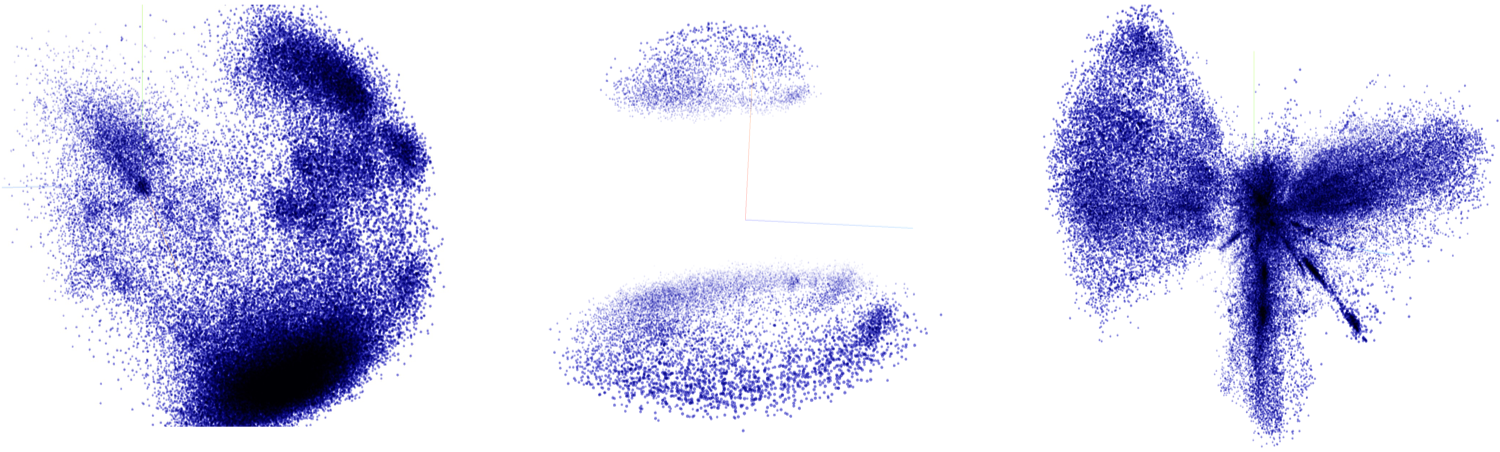}
\caption{Visualization of the $\ourGraph$ graph embeddings ($d$=500) learned by Node2vec (top left), Metapath2vec (middle) and Poincar\'e (right). The shape of the visualizations demonstrate the distinct method objective and embedding characteristics (Node2vec: neighbourhood correlations; Metapath2vec: distinct node types; Poincare: hierarchical relations)}
\label{fig:emb_viz}
\end{centering}
\vspace{-4mm}
\end{figure*}



\subsection{Evaluation tasks}
We use a suite of knowledge graph-based healthcare application tasks to evaluate the learned concept representations. 

\textbf {Graph Quality Evaluation Tasks}
\begin{enumerate}
    \item \textbf{Multi Label Classification}: We test the accuracy of learned embeddings in capturing a node's type. Each node is assigned a semantic type from a finite set $L$, and a classifier model is trained on the partial set of node embeddings to predict the labels for remaining nodes in the test set.  
    \item \textbf{Link prediction}: We test the accuracy of learned embeddings in identifying the presence or absence of an edge in the  knowledge graph. A classifier model is trained on the cosine similarity of the embeddings to predict the existence of a relation between concept pairs.  
\end{enumerate}

\textbf {Domain Tasks}  
\begin{enumerate}
\item \textbf{Concept similarity}: We follow the benchmark generation strategy and cosine similarity measure outlined by \cite{beam2018clinical, choi2016multi} based on statistical power to evaluate ``relatedness'' in similar concepts. For a given relationship
$(\mathbf{x},\mathbf{y})$, a null distribution of cosine similarities is computed using bootstrapping samples $(\mathbf{x}^*,\mathbf{y}^*)$, where $\mathbf{x}^*$ and $\mathbf{y}^*$ belong to the same category as $\mathbf{x}$ and $\mathbf{y}$, respectively. Then, the statistical power $p$ of the bootstrap distribution is reported to reject the null hypothesis (i.e., no relationship).

\begin{figure}[htbp]
\begin{centering}
\includegraphics[width=1.0\columnwidth]{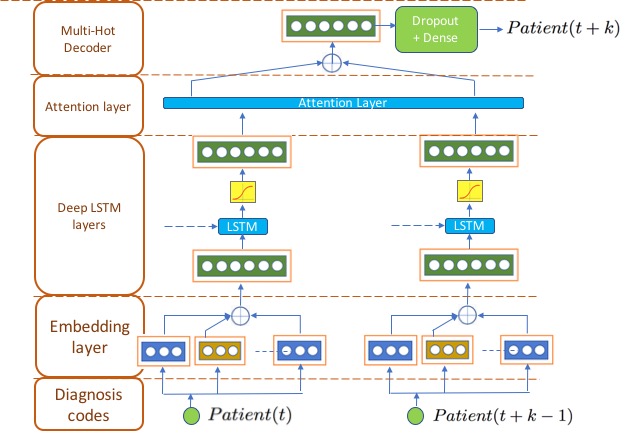}
\caption{Architecture of the deep learning model to predict diagnostic codes from past EHR information}
\label{fig:prediction_architecture}
\end{centering}
\vspace{-4mm}
\end{figure}

\item \textbf{Patient state prediction} 
Given a set of diagnosis codes for $k$ visits, $C_1$, $C_2$, ..., $C_{t}$, we predict the set of diagnosis codes for the $(t+1)$ visit, $C_{t+1}$. Following \cite{choi2017gram}, we train a long short term memory (LSTM) model using cross-entropy loss function to capture patient state transitions over several visits (Fig. \ref{fig:prediction_architecture}).
 Given $\vbb_1, \vbb_2, \vbb_3, \cdots, \vbb_t$ as vector representation of patient states for $t$ timesteps and $\textbf{W}$ representing the weights of the single LSTM layer, equation \ref{eq:lstm1} and \ref{eq:lstm2} describe our prediction task.   
\begin{equation} \label{eq:lstm1}
    \widehat{\yb}_{t} = \widehat{\xb}_{t+1}  = \mathrm{Softmax}(\textbf{W} \hb_{t} + \bb)
\end{equation}
\vspace{-4mm}
\begin{equation} \label{eq:lstm2}
    \hb_1, \hb_2, \ldots, \hb_{t} = \textrm{LSTM}(\vbb_1, \vbb_2, \ldots, \vbb_{t}; \theta_{r})
\end{equation}
 We characterize the patient state prediction as a set of three different objective functions: 1) Predict all diagnosis codes for the next visit, 2) Predict $top\_k$  frequent diagnosis, and 3) Predict $top\_k$ rare diagnosis (occurred in at least in 100 visits).


\end{enumerate}


\section{Experiments}

\subsection{Dataset description}
\label{dataset}

We start from the UMLS semantic network \cite{mccray1989umls} and select the subset of clinical concepts of relevance to patient level modeling, limiting to the following semantic groups : Anatomy (ANAT), Chemicals \& Drugs (CHEM), Disorders (DISO), and Procedures (PROC).\footnote{The list of selected concepts is available from \url{https://anonymous.4open.science/r/0651fc32-8eff-4454-b537-5c00bb12ea19/}.} We used MRSTY.RRF, MRCONSO.RRF, and MRREL.RRF tables from the UMLS for this purpose. MRSTY.RRF was used to define the semantic type of extracted  concepts. 
Finally, MRREL.RRF provides the relations between the selected concepts. We subsequently refer to the extracted graph as ${\ourGraph}$. Figure \ref{fig:snomed} shows the number of concepts in each  concept group and the number of unique relation types between them. 


\begin{table*}[h!]
\centering
\begin{tabular}{rrrrrr}
  \hline
  & \textbf{Node2vec} & \textbf{Metapath2vec} & \textbf{Poincare} & \textbf{CUI2vec} & \textbf{Med2vec}  \\ 
  \hline
 \textbf{Node Classification} & 0.817 & 0.3287 & \textbf{0.8579} & 0.5685 & 0.0409 \\ 
 \textbf{Link Prediction} & \textbf{0.986} & 0.3988 & 0.7135 & 0.7222 & 0.8665 \\ 
 \textbf{Concept Similarity} (D1) & \textbf{0.79} & 0.3 & 0.7 & 0.16  & NA \\
 \textbf{Concept Similarity} (D3) & \textbf{0.90} &0.46  &0.31  & 0.15 & NA \\
 \textbf{Concept Similarity} (D5) & \textbf{0.81} & -0.32 & -0.06  & -0.01 & NA \\
 \textbf{Patient State Prediction (All Diagnosis)} & 0.3938 & 0.3359 & \textbf{0.4197} & 0.3948 & 0.3881  \\
 \textbf{Patient State Prediction (Frequent 20)} & 0.8465 & \textbf{0.9749} & 0.85 & 0.8035 & 0.7980 \\
 \textbf{Patient State Prediction (Rare 20)} & 0.018 & 0.001 & \textbf{0.019} & \textbf{0.019}  & 0.011 \\
   \hline
\end{tabular}
\caption{Performance evaluation of embeddings on each task. Best performing method is highlighted for each task (using data from best performing embedding size for each method). Evaluation results show that knowledge graph-based embeddings outperform the state of art (Med2vec and CUI2vec) on all tasks.}
\label{tab:results}
\end{table*}

\subsection{Experimental setup for embedding learning}

We generated embeddings using the three methods (Fig. \ref{fig:emb_viz}) (Node2vec, Metapath2vec and Poincar\`e) for dimensions (20, 50, 100, 200, 500).

\textbf{Node2vec} For Node2vec\cite{grover2016node2vec}, we varied the number of walks per node between $W = (5, 10, 20)$.  Random walk length was varied between $L = (5, 20, 40, 80)$. Other internal parameters, such as batch size and epochs, were set to default values. The best results were obtained at $W=10, L=20$.

\textbf{Metapath2vec} embeddings require specification and sampling specific instances of desired path patterns or ``meta-paths" in the graph\cite{dong2017metapath2vec}. We used patterns that connect nodes in disease category to drugs and set random walk parameters to W=20 and L = 5.

\textbf{Poincar\'e} We computed Poincar\'e embeddings using an implementation of \cite{nickel2017poincare} included in the gensim software \cite{rehurek_lrec} framework.  The embeddings were trained with L2 regularization for 50 epochs using a learning rate of 0.1 and a burn-in parameter of 10. A non-zero burn-in value initializes the vectors to a uniform distribution and trains with a much reduced learning rate for a few epochs. This yields a better initial angular layout of the vectors and increases the robustness of final vectors to random bad initialization. 

\textbf{CUI2Vec}: CUI2vec embeddings \cite{beam2018clinical} were made available by the corresponding authors for the purpose of this evaluation. We used 500-dimensional CUI2vec embeddings, covering nearly 108K concepts in the UMLS thesaurus. We used the UMLS mapping of CUI concepts to snomed concept ids (SCUI) for evaluation purposes.

\textbf{Med2vec} : The med2vec \cite{choi2016multi} embeddings are 200-dimensional embeddings available for 27523 ICD\_9 concepts. We use the ICD to SNOMED-CT concept map  made available by UMLS for mapping ICD\_9 to SNOMED concept ids where needed for evaluation.

\begin{figure*}[h!]
\begin{centering}
\includegraphics[width=2.0\columnwidth]{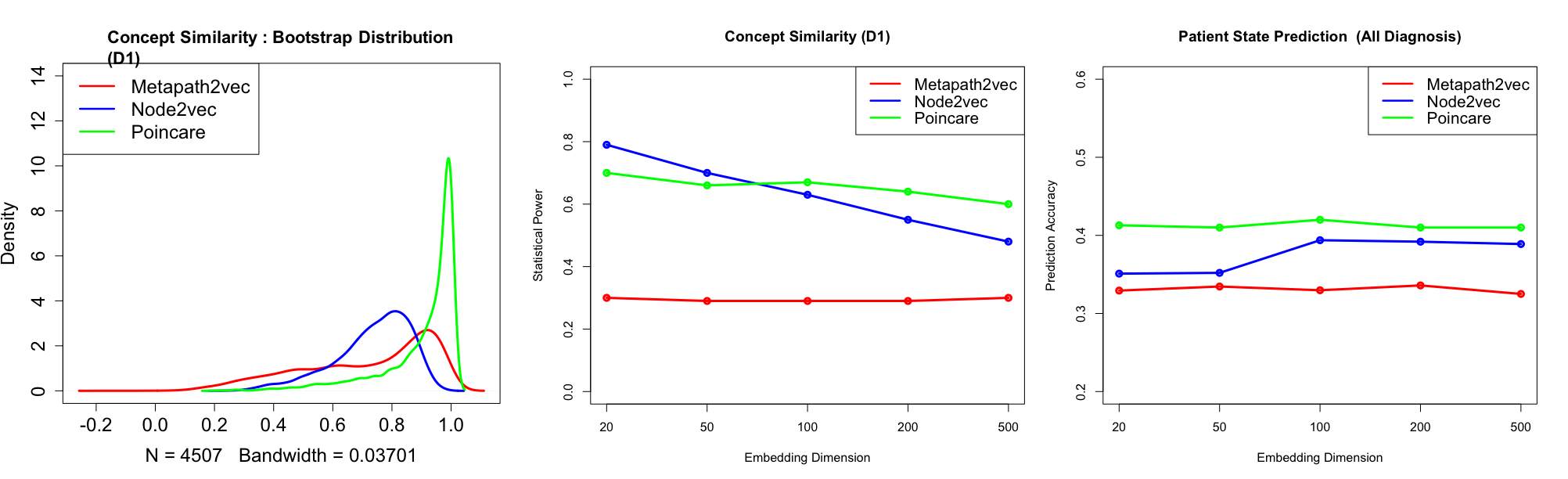}
\caption{Left to right: (a) Bootstrap distribution of cosine similarity for each method on D$_1$, where the density is calculated using the \textit{density} function from the R programming language, and the bandwidth is the standard deviation of the kernel. The narrower curve means more statistical power between the concept pairs. (b) Impact of embedding dimension $d$ on concept similarity and (c) Impact of embedding dimension $d$ on patient state prediction}
\label{fig:emb_dims}
\end{centering}
\vspace{-4mm}
\end{figure*}

\section{Results}
Table \ref{tab:results} shows the results for each embedding learning method along with CUI2vec and Med2vec on the evaluation tasks.


\textbf{Multi Label Classification}
We use the semantic types provided in the UMLS graph (MRSTY.RRF datafile) for each node as its class label; $\ourGraph$ had ($|L| = 43$) unique labels, making it a particularly challenging task. We calculate accuracy as the fraction of nodes for which the predicted label matched the ground truth. We observe that Poincar\'e yielded the best accuracy ($d=500$), outperforming CUI2vec by 50\% and Med2vec by several orders of magnitude. As Poincar\'e primarily samples a node\'s type hierarchy for learning, it's highly suited for healthcare applications that need embeddings highly cognizant of node type \cite{choi2017gram}.

\textbf{Link Prediction}
We use a simple linear SVM classifier model for testing the link prediction ability of learned embeddings.  
For training, we sampled about 2\% of a total 4.7 million relations available in $\ourGraph$ to restrict the training time to few hours. We further constructed a negative sample (i.e., links that are not present in the graph) of the same count similar to ~\cite{zhang2018link}. The accuracy is calculated as the fraction of $|correct predictions|/|test pairs|$. We observe Node2vec captures maximal correlation between distant concepts, outperforming Med2vec by 15\% and CUI2vec by 36\% in the link prediction task.  
\textbf{Concept similarity}
We create five different datasets from  ${\ourGraph}$  to capture the unique bootstrap distributions between concept groups. D$_1$ and D$_2$ represent hierarchical ('isa') relations for DISO and CHEM concept groups respectively; while D$_3$ and D$_4$ used all non hierarchical relations for DISO and CHEM groups, representing relatedness between concepts in same semantic group. D$_5$ contains ``all'' relations between DISO and CHEM, representing relatedness across semantic groups. Table \ref{tab:results} shows statistical power for each method.
Med2vec is excluded as it did not have enough concept pair samples for any dataset. 
Poincar\'e and Node2vec perform well for D$_1$ and D$_2$ (Figure \ref{fig:emb_dims}(a)).
However, when the hierarchical relationships are excluded from the benchmarking (D$_3$, D$_4$, D$_5$), the Node2vec method provides larger statistical power, showing 5-6x improvement over CUI2vec.

\textbf{Patient state prediction}
The patient model is trained on the openly available MIMIC-III \cite{mimic} EHR dataset, which contains ICU visit records for more than than 40,000 patients over 11 years.  Each visit of the patient is represented as a collection of ICD\_9 diagnosis code, which are mapped to the coarser grained 284 dimensions using a well-organized ontology called Clinical Classifications Software (CCS)\cite{donnelly2006snomed}. Table \ref{tab:results} shows the results for all three prediction tasks. 
We observe that Poincar\'e-based embeddings outperform other methods for prediction of all diagnosis, showing 6\% improvement over CUI2vec  (next best performing method). CUI2vec  performs equally well as Poincar\'e on the prediction of rare diagnoses, while Metapath2vec outperforms in predicting the most frequent diagnosis, improving 20\% over CUI2vec and Med2vec. Our results are in line with recent work by Choi et al \cite{choi2017gram}, which showed that incorporating clinical concept hierarchies improved patient models significantly. Our Poincar\'e embeddings capture these hierarchical relations and hence are best suited for downstream clinical tasks.

\textbf{Impact of embedding dimensions}: Figure \ref{fig:emb_dims} (b) and (c) shows the accuracy of embeddings for different dimensions $d$ for concept similarity and patient model respectively. Metapath2vec remains invariable to dimension size, while Poincar\'e and Node2vec exhibit best performance for $d$=100 for both tasks. The results for other evaluation tasks followed similar trends but have been omitted for brevity. 


\vspace{-4mm}
\section{Conclusions}
We propose and develop a workflow for learning medical concept representations for a healthcare knowledge graph that is widely-used for building clinical decision support tools in medical practice and in research applications. We evaluate and demonstrate that knowledge graph-driven embedding outperforms state-of-the-art in several healthcare applications. Our analysis suggests that Poincar\'e-based embeddings learned from hierarchical relations are highly efficient for patient state prediction models and in capturing node type (classification). In addition, Node2vec embeddings are best suited for capturing ``relatedness'' or concept similarity. 

\bibliography{snomed_embedding}

\begin{thebibliography}{15}
\providecommand{\natexlab}[1]{#1}
\providecommand{\url}[1]{\texttt{#1}}
\expandafter\ifx\csname urlstyle\endcsname\relax
  \providecommand{\doi}[1]{doi: #1}\else
  \providecommand{\doi}{doi: \begingroup \urlstyle{rm}\Url}\fi

\bibitem[Beam et~al.(2018)Beam, Kompa, Fried, Palmer, Shi, Cai, and
  Kohane]{beam2018clinical}
A~L Beam, B~Kompa, I~Fried, N~P Palmer, X~Shi, T~Cai, and I~S Kohane.
\newblock Clinical concept embeddings learned from massive sources of medical
  data.
\newblock \emph{arXiv preprint arXiv:1804.01486}, 2018.

\bibitem[Choi et~al.(2016)Choi, Bahadori, Searles, Coffey, Thompson, Bost,
  Tejedor-Sojo, and Sun]{choi2016multi}
Edward Choi, Mohammad~Taha Bahadori, Elizabeth Searles, Catherine Coffey,
  Michael Thompson, James Bost, Javier Tejedor-Sojo, and Jimeng Sun.
\newblock Multi-layer representation learning for medical concepts.
\newblock In \emph{Proceedings of the 22nd ACM SIGKDD International Conference
  on Knowledge Discovery and Data Mining}, pp.\  1495--1504. ACM, 2016.

\bibitem[Choi et~al.(2017)Choi, Bahadori, Song, Stewart, and Sun]{choi2017gram}
Edward Choi, Mohammad~Taha Bahadori, Le~Song, Walter~F Stewart, and Jimeng Sun.
\newblock Gram: graph-based attention model for healthcare representation
  learning.
\newblock In \emph{Proceedings of the 23rd ACM SIGKDD International Conference
  on Knowledge Discovery and Data Mining}, pp.\  787--795. ACM, 2017.

\bibitem[De~Vine et~al.(2014)De~Vine, Zuccon, Koopman, Sitbon, and
  Bruza]{devine2014med}
Lance De~Vine, Guido Zuccon, Bevan Koopman, Laurianne Sitbon, and Peter Bruza.
\newblock Medical semantic similarity with a neural language model.
\newblock In \emph{Proceedings of the 23rd ACM international conference on
  conference on information and knowledge management}, pp.\  1819--1822. ACM,
  2014.

\bibitem[Dhingra et~al.(2018)Dhingra, Shallue, Norouzi, Dai, and
  Dahl]{dhingra2018embedding}
B~Dhingra, C~J Shallue, M~Norouzi, A~M Dai, and G~E Dahl.
\newblock Embedding text in hyperbolic spaces.
\newblock \emph{arXiv preprint arXiv:1806.04313}, 2018.

\bibitem[Dong et~al.(2017)Dong, Chawla, and Swami]{dong2017metapath2vec}
Yuxiao Dong, Nitesh~V Chawla, and Ananthram Swami.
\newblock metapath2vec: Scalable representation learning for heterogeneous
  networks.
\newblock In \emph{Proceedings of the 23rd ACM SIGKDD international conference
  on knowledge discovery and data mining}, pp.\  135--144. ACM, 2017.

\bibitem[Donnelly(2006)]{donnelly2006snomed}
K~Donnelly.
\newblock {SNOMED-CT}: The advanced terminology and coding system for
  e{H}ealth.
\newblock \emph{Studies in health technology and informatics}, 121:\penalty0
  279, 2006.

\bibitem[Grover \& Leskovec(2016)Grover and Leskovec]{grover2016node2vec}
Aditya Grover and Jure Leskovec.
\newblock node2vec: Scalable feature learning for networks.
\newblock In \emph{Proceedings of the 22nd ACM SIGKDD international conference
  on Knowledge discovery and data mining}, pp.\  855--864. ACM, 2016.

\bibitem[Johnson~AEW \& RG(2016)Johnson~AEW and RG]{mimic}
Shen L Lehman L Feng M Ghassemi M Moody B Szolovits P Celi~LA Johnson~AEW,
  Pollard~TJ and Mark RG.
\newblock {MIMIC-III}, a freely accessible critical care database.
\newblock \emph{Scientific Data (2016). DOI: 10.1038/sdata.2016.35. Available
  at: http://www.nature.com/articles/sdata201635}, 2016.

\bibitem[McCray(1989)]{mccray1989umls}
Alexa~T McCray.
\newblock The {UMLS} semantic network.
\newblock In \emph{Proceedings. Symposium on Computer Applications in Medical
  Care}, pp.\  503--507. American Medical Informatics Association, 1989.

\bibitem[Mi{\~{n}}arro{-}Gim{\'{e}}nez
  et~al.(2015)Mi{\~{n}}arro{-}Gim{\'{e}}nez, Mar{\'{\i}}n{-}Alonso, and
  Samwald]{minarro2015}
J~A Mi{\~{n}}arro{-}Gim{\'{e}}nez, O~Mar{\'{\i}}n{-}Alonso, and M~Samwald.
\newblock Applying deep learning techniques on medical corpora from the {W}orld
  {W}ide {W}eb: a prototypical system and evaluation.
\newblock \emph{CoRR}, abs/1502.03682, 2015.
\newblock URL \url{http://arxiv.org/abs/1502.03682}.

\bibitem[Nickel \& Kiela(2017)Nickel and Kiela]{nickel2017poincare}
Maximillian Nickel and Douwe Kiela.
\newblock Poincar{\'e} embeddings for learning hierarchical representations.
\newblock In \emph{Advances in neural information processing systems}, pp.\
  6338--6347, 2017.

\bibitem[Perozzi et~al.(2014)Perozzi, Al-Rfou, and Skiena]{perozzi2014deepwalk}
Bryan Perozzi, Rami Al-Rfou, and Steven Skiena.
\newblock Deepwalk: Online learning of social representations.
\newblock In \emph{Proceedings of the 20th ACM SIGKDD international conference
  on Knowledge discovery and data mining}, pp.\  701--710. ACM, 2014.

\bibitem[{\v R}eh{\r u}{\v r}ek \& Sojka(2010){\v R}eh{\r u}{\v r}ek and
  Sojka]{rehurek_lrec}
R~{\v R}eh{\r u}{\v r}ek and P~Sojka.
\newblock {Software Framework for Topic Modelling with Large Corpora}.
\newblock In \emph{{LREC Workshop on New Challenges for NLP Frameworks}}, 2010.

\bibitem[Zhang \& Chen(2018)Zhang and Chen]{zhang2018link}
Muhan Zhang and Yixin Chen.
\newblock Link prediction based on graph neural networks.
\newblock In \emph{Advances in Neural Information Processing Systems}, pp.\
  5165--5175, 2018.

\end{thebibliography}
\bibliographystyle{iclr2019_conference}

\end{document}